\documentclass[sigconf]{aamas}  
\usepackage{balance}  

\usepackage{booktabs}
\usepackage{hyperref}
\usepackage{graphicx}
\usepackage{algorithm,algpseudocode}
\usepackage{algorithmicx}
\usepackage{float}
\usepackage{paralist}
\usepackage{amsmath}
\usepackage{breqn}
\usepackage{cite}
\usepackage{subfig}
\usepackage{multicol}
\usepackage{comment}

\DeclareMathOperator{\dom}{dom}
\DeclareMathOperator*{\argmax}{argmax}

\setcopyright{ifaamas}  
\acmDOI{}  
\acmISBN{}  
\acmConference[AAMAS'19]{Proc.\@ of the 18th International Conference on Autonomous Agents and Multiagent Systems (AAMAS 2019)}{May 13--17, 2019}{Montreal, Canada}{N.~Agmon, M.~E.~Taylor, E.~Elkind, M.~Veloso (eds.)}  
\acmYear{2019}  
\copyrightyear{2019}  
\acmPrice{}  

\renewcommand\footnotetextcopyrightpermission[1]{} 
\begin{document}

\title{Urban Driving with Multi-Objective Deep Reinforcement Learning}  


\author{Changjian Li}
\affiliation{
  \institution{University of Waterloo}
  \streetaddress{200 University Ave. W.}
  \city{Waterloo} 
  \state{Ontario}
  \country{Canada}
  \postcode{N2L 3G1}
}
\email{changjian.li@uwaterloo.ca}

\author{Krzysztof Czarnecki}
\affiliation{
  \institution{University of Waterloo}
  \streetaddress{200 University Ave. W.}
  \city{Waterloo} 
  \state{Ontario}
  \country{Canada}
  \postcode{N2L 3G1}
}
\email{kczarnec@gsd.uwaterloo.ca}

\begin{abstract}  
Autonomous driving is a challenging domain that entails multiple aspects: a vehicle should be able to drive to its destination as fast as possible while avoiding collision, obeying traffic rules and ensuring the comfort of passengers. In this paper, we present a deep learning variant of \emph{thresholded lexicographic Q-learning} for the task of urban driving. Our multi-objective DQN agent learns to drive on multi-lane roads and intersections, yielding and changing lanes according to traffic rules. We also propose an extension for \emph{factored Markov Decision Processes} to the DQN architecture that provides auxiliary features for the Q function. This is shown to significantly improve data efficiency.~\footnote{Data efficiency as measured by the number of training steps required to achieve similar performance.} We then show that the learned policy is able to zero-shot transfer to a ring road without sacrificing performance.
\end{abstract}

%


\maketitle


\section{Introduction}
Deep reinforcement learning (DRL)~\citep{DBLP:journals/corr/MnihKSGAWR13} has seen some success in complex tasks with large state space. However, the task of autonomous urban driving remains challenging, partly due to the many aspects involved: the vehicle is not only expected to avoid collisions with dynamic objects, but also follow all the traffic rules and ensure the comfort of passengers. One motivation for the multi-objective approach is the difficulty in designing a scalar reward that properly weighs the importance of each aspect of driving so that the designer's original intention is reflected. Although there have been attempts to learn the reward function through inverse reinforcement learning~\citep{DBLP:conf/icml/PieterN04, DBLP:conf/aaai/ZiebartMBD08}, these methods add additional computational expenses, and require the availability of demonstrations. Another motivation comes from the problem of exploration~\citep{DBLP:conf/nips/BellemareSOSSM16, DBLP:conf/nips/OsbandBPR16}. If the agent explores randomly, it might hardly have the chance of reaching the intersection, so the traffic rules at the intersection might never be learned. In contrast, if each aspect is learned separately, we would have the flexibility of choosing which aspect to explore in a given state. In this paper, we consider a multi-objective RL approach to the problem of urban driving, where each objective is learned by a separate agent. These agents collectively form a combined policy that takes all these objectives into account. In addition to those mentioned above, there are several advantages of the multi-objective approach:
\begin{inparaenum}
\item Since each objective only considers one aspect of driving, the entire task is divided into smaller, simpler tasks, e.g., smaller state space can be used for each objective.
\item In a new task where only some of the objectives change, the learned policy for other objectives can be reused or transferred.
\item For some of the objectives, the desired behavior might be easy to specify manually. The multi-objective architecture allows these behaviors to be implemented with rule-based systems without creating integration issues.
\end{inparaenum}

In this paper, we adopt the thresholded lexicographic Q-learning framework proposed by \citeauthor{DBLP:conf/icml/GaborKS98}~\citep{DBLP:conf/icml/GaborKS98}, and adapt it to the deep learning setting. We set an adaptive threshold for the Q value of each objective, and at each state, the set of admissible actions is restricted by lexicographically applying the threshold to each objective. Therefore, the policy obtained either satisfies all constraints, or, if that's not possible, satisfies the constraints for the more important objectives. We believe that this paradigm is similar to how human drivers drive, e.g., a human driver would aim at guaranteeing safety before considering other aspects such as traffic rules and comfort.

Most existing RL approaches for autonomous driving consider a state space of either raw visual/sensor input~\citep{DBLP:journals/corr/SallabAPY16, DBLP:journals/corr/IseleCSF17}, or the kinematics of a few immediately surrounding vehicles~\citep{DBLP:conf/itsc/NgaiY07, DBLP:journals/corr/abs-1803-09203}. Since road and lane information is not explicitly considered, the policy learned using these types of state space in limited scenarios cannot be expected to be transferable to roads with different geometry. In this work, we design a hybrid (of continuous and discrete) state space that not only includes the state of surrounding vehicles, but also geometry-independent road topology information. We show that the policy trained in a four-way intersection using the proposed state space can be zero-shot transferred to a ring road.

As more surrounding vehicles are considered, the complexity of the problem increases exponentially. However, a human learner is able to reduce complexity by focusing on a few important vehicles in a particular situation, presumably because a human learner exploits some sort of structure of the problem. For example, if a human learner is following a car too close, he not only knows the fact that he \textit{is} following too close, but he also knows: 
\begin{inparaenum}
  \item \textit{which car} he is following too close; and
  \item the car on the other side of intersection has very little, if anything, to do with the situation.
\end{inparaenum}
In other words, in addition to viewing the state space as a whole, humans are, at the same time, learning on each individual factored state space (the car ahead, and the car on the other side of intersection, etc.) as well, then they use the knowledge learned on the factored state space to help with the original task. To mimic this behaviour, we propose to decompose \emph{factored MDPs} into auxiliary tasks, then the factored Q functions learned on the factored state space can be used as additional features for the original Q function. This is shown to significantly improve data efficiency.

\section{Related Works}
There has been emerging research on autonomous driving using reinforcement learning. \citeauthor{DBLP:journals/corr/SallabAPY16}~\citep{DBLP:journals/corr/SallabAPY16} compared two DRL-based  end-to-end driving models, DQN and DDPG,  in TORCS car racing simulator. \citeauthor{DBLP:journals/corr/abs-1803-09203}~\citep{DBLP:journals/corr/abs-1803-09203} proposed to represent the Q function as a quadratic function to deal with continuous action space. \citeauthor{DBLP:conf/itsc/NgaiY07} attempted multi-objective reinforcement learning for learning takeover maneuver~\citep{DBLP:conf/itsc/NgaiY07}, where they scalarized the learned Q functions of each objective by weighted sum to form a single policy. The sensor input was quantized into a discrete state space, and tabular Q-learning was used. 

\citeauthor{DBLP:journals/corr/IseleCSF17}~\citep{DBLP:journals/corr/IseleCSF17} trained a DQN agent that can navigate through an intersection, where the scenario is somewhat similar to ours. However, our scenario is much more complex due to several major differences:
\begin{inparaenum}
\item Their agent is trained in \emph{one single} scenario each time, and the evaluation is done in the same scenario. Our agent is trained in random scenarios within the map, and evaluated in random scenarios.
\item Their agent only deals with the intersection part. The agent starts at one side of the intersection and all it does is to find a gap to enter, the episode ends as soon as it reaches the other side of the intersection. Our agent needs to drive safely to the intersection and if necessary, slow down, yield and change lanes before entering the intersection.
\item No traffic rules are considered in their work.
\end{inparaenum}

There have also been research that exploits the temporal structure of the policy (hierarchical RL~\citep{DBLP:conf/nips/ParrR97, DBLP:journals/ai/SuttonPS99, DBLP:journals/jair/Dietterich00}). \citeauthor{DBLP:conf/iros/PaxtonRHK17}~\citep{DBLP:conf/iros/PaxtonRHK17} proposed a method where they designed a set of high level \emph{options}~\citep{DBLP:journals/ai/SuttonPS99}, and used Monte Carlo tree search and DDPG \citep{DBLP:journals/corr/LillicrapHPHETS15} to learn the high level and low level policy, respectively.

The idea of thresholded lexicographic ordering can be traced back to the \textit{hierarchical optimization criteria} proposed by \citeauthor{waltz1967engineering}~\citep{waltz1967engineering} for multi-objective optimization. \citeauthor{DBLP:conf/icml/GaborKS98}~\citep{DBLP:conf/icml/GaborKS98} extended this idea to reinforcement learning. \citeauthor{DBLP:conf/aaai/WrayZM15}~\citep{DBLP:conf/aaai/WrayZM15} and \citeauthor{DBLP:conf/aaaifs/PinedaWZ15}~\citep{DBLP:conf/aaaifs/PinedaWZ15} considered an adaptive threshold that depends on the learned value function.

The factored Q functions used in this paper can be thought of as a form of generalized value function proposed by \citeauthor{DBLP:conf/atal/SuttonMDDPWP11}~\citep{DBLP:conf/atal/SuttonMDDPWP11}. The auxiliary POMDPs used to train the factored Q functions can be considered a form of unsupervised auxiliary tasks introduced by  \citeauthor{DBLP:journals/corr/JaderbergMCSLSK16}~\citep{DBLP:journals/corr/JaderbergMCSLSK16}. The difference is that we exploit the structure of factored MDP and factorize the \emph{original task itself} into auxiliary tasks, then the learned factored Q functions are used directly as features (basis functions) for the original Q function.

An overview of multi-objective optmization can be found in \citep{Marler2004}, and a summary of multi-objective sequential decision making is given in \citep{DBLP:journals/corr/RoijersVWD14}.

\section{Background}
\subsection{Multi-Objective Reinforcement Learning}
Multi-objective reinforcement learning (MORL) is concerned with multi-objective Markov decision processes (MOMDPs) $(S, A, P, \mathbf{r}, \gamma)$ , where $S$ is a finite set of states; $A$ is a finite set of actions; $P(\mathbf{s'}|\mathbf{s}, a)$ is the transition probability from state $\mathbf{s}$ to state $\mathbf{s'}$ taking action $a$; $\mathbf{r}(\mathbf{s}, a) = [r_{1}(\mathbf{s}, a), r_{2}(\mathbf{s}, a),... , r_{k}(\mathbf{s}, a)]$ and $\gamma = [\gamma_{1}, \gamma_{2},... , \gamma_{k}]$ are the rewards and discount factors for the $k$ objectives, respectively.~\footnote{$\mathbf{r}(\mathbf{s}, a)$ can be generalized to $\mathbf{r}(\mathbf{s}, a, \mathbf{s'})$.} Fixing some enumeration from 1 to $|S|$ of the finite state space $S$, we denote $\mathbf{v}_{i}^{\pi}$ as a column vector whose $j$th element is the value function of the $i$th objective evaluated at the $j$th state $\mathbf{s}^{j}$. Precisely:
\begin{displaymath}
\mathbf{v}_{i}^{\pi} = [v_{i}^{\pi}(\mathbf{s}^{1}), v_{i}^{\pi}(\mathbf{s}^{2}), ..., v_{i}^{\pi}(\mathbf{s}^{|S|})]^T
\end{displaymath}
\begin{displaymath}
v_{i}^{\pi}(\mathbf{s}) = E[\sum_{t=0}^{t=\infty}{\gamma_{i}^{t}r_{i}(\mathbf{s}_{t}, a_{t})} | \pi, \mathbf{s}], \forall \mathbf{s} \in S, i =1, 2,..., k
\end{displaymath}
MORL aims to find some policy $\pi: S \to A$, such that
\begin{displaymath}
\pi(s) = \mbox{argmax}_{\pi} \mathbf{V}^{\pi}
\end{displaymath}
\begin{displaymath}
\mathbf{V}^{\pi} = [\mathbf{v}_{1}^{\pi}, \mathbf{v}_{2}^{\pi},... \mathbf{v}_{k}^{\pi}]
\end{displaymath}
Different definition of order relation on the \emph{feasible criterion space} $\{\mathbf{v} | \pi \in {S \to A}\}$ leads to different MORL algorithms, or in other words, different MORL algorithm implicitly defines such an (partial or total) order relation. In this paper, we consider a thresholded lexicographic approach that we deem suitable for autonomous driving. 

\subsection{Thresholded Lexicographic Q-learning}
Assuming lexicographic ordering $1, 2, ..., k$ on the $k$ objectives of MOMDP $(S, A, T, \mathbf{r}, \gamma)$, and $\tau_{i}$ a $local$ threshold that specifies the minimum admissible value for each objective, thresholded lexicographic Q-learning finds k sets of policies $\Pi_{i}, i = 1, 2, ..., k$ that maximize 
$\{\hat{Q}_{1}^{*}(\mathbf{s}, a), \hat{Q}_{2}^{*}(\mathbf{s}, a), ..., \hat{Q}_{i}^{*}(\mathbf{s}, a)\}$
in lexicographic order:
\begin{align}
\label{eq:TLQ_policy}
\begin{split}
\Pi_i \stackrel{def}{=} & \bigg\{\pi_{i} \in \Pi_{i-1} \bigg| \pi_{i}(\mathbf{s}) = \mbox{argmax}_{a \in \{\pi_{i-1}(\mathbf{s})| \pi_{i-1} \in \Pi_{i-1}\}} \hat{Q}_{i}^{*}(\mathbf{s}, a) \\ 
                        & \quad , \forall \mathbf{s} \in S \bigg\}, i = 1, 2, ..., k
\end{split}
\end{align}
with $\Pi_{0}$ being the set of all deterministic stationary policies, and $\hat{Q}_{i}^{*}(\mathbf{s}, a)$ is the Q function rectified to $\tau_{i}$:
\begin{equation}
\hat{Q}_{i}^{*}(\mathbf{s}, a) \stackrel{def}{=} \min(\tau_{i}, Q_{i}^{*}(\mathbf{s}, a))
\end{equation}
Here, $Q_{i}^{*}(\mathbf{s}, a)$ is the maximum expected accumulative reward \emph{over all policies $\pi_{i-1} \in \Pi_{i-1}$} starting from state $\mathbf{s}$ and action $a$. It follows that 
\begin{align}
\label{Qhat}
\begin{split}
& \hat{Q}_{i}^{*}(\mathbf{s}, a) = \\
& \min {\left( \tau_{i}, \quad r_{i}(\mathbf{s},a) + 
    \gamma_{i} \sum_{\mathbf{s}'}{P(\mathbf{s}'|\mathbf{s},a) \max_{a' \in \{\pi_{i-1}(\mathbf{s})| \pi_{i-1} \in \Pi_{i-1}\}}{Q_{i}^{*}(\mathbf{s}', a')}} \right)} \geq \\
& \min {\left( \tau_{i}, \quad r_{i}(\mathbf{s},a) + 
    \gamma_{i} \sum_{\mathbf{s}'}{P(\mathbf{s}'|\mathbf{s},a) \max_{a' \in \{\pi_{i-1}(\mathbf{s})| \pi_{i-1} \in \Pi_{i-1}\}}{\hat{Q}_{i}^{*}(\mathbf{s}', a')}} \right)}
\end{split}
\end{align}
\citeauthor{DBLP:conf/icml/GaborKS98}~\citep{DBLP:conf/icml/GaborKS98} propose to approximate $\hat{Q}_{i}^{*}(\mathbf{s}, a)$ by treating the inequality in Eq. \ref{Qhat} as equality, and do the following value iteration:
\begin{align}
\label{Qhat_eq}
\begin{split}
& \hat{Q}_{i}^{*}(\mathbf{s}, a)  := \\
& \min {\left( \tau_{i}, \quad r_{i}(\mathbf{s},a) + 
    \gamma_{i} \sum_{\mathbf{s}'}{P(\mathbf{s}'|\mathbf{s},a) \max_{a' \in \{\pi_{i-1}(\mathbf{s})| \pi_{i-1} \in \Pi_{i-1}\}}{\hat{Q}_{i}^{*}(\mathbf{s}', a')}} \right)}
\end{split}
\end{align}

Unfortunately, following a policy $\pi \in \Pi_{i}$ as in Eq. \ref{eq:TLQ_policy} doesn't guarantee that $v_{i}^{\pi}(\mathbf{s}) > \tau_{i}$, even if $\exists \pi' \in \Pi_{i-1}, v_{i}^{\pi'}(\mathbf{s}) > \tau_{i}, \forall \mathbf{s} \in S$. In fact, the policies in $\Pi_{i}$ can be arbitrarily bad. Therefore the algorithm is only appropriate for problems that are tolerant to small \emph{local} imperfections of the policy, which needs to be kept in mind when designing the reward function. 

\subsection{Factored MDP}
In almost all real-life domains, the state space of MDPs is very large. Fortunately,  many large MDPs have some internal structures. The factored MDP framework aims to exploit these internal structures by representing a state $s \in S$ with a set of  \emph{state variables} $s = (s_{1}, s_{2}, ..., s_{m})$. A \emph{locally-scoped} function is defined as a function that depends only on a subset of the state variables~\citep{DBLP:journals/jair/GuestrinKPV03}, and the subset is called the \emph{scope} of the function. There are mainly two types of structures that are covered in literature: the \emph{additive} structure and \emph{context-specific} structure. As an example of context-specific structure, consider the driving domain: the traffic signal is not affected by the current speed of the cars travelling at the intersection. Moreover, reward function might be a sum of a few locally-scoped rewards, such as the (negative) reward for collision and the reward for maintaining steady speed, which is an example of additive structure. In this paper, a method to exploit these structures in DQN is proposed.

\section{Approach}

\subsection{Thresholded Lexicographic DQN}
Approximating $\hat{Q}_{i}^{*}(\mathbf{s}, a)$ directly by Eq. \ref{Qhat_eq} as proposed by \citeauthor{DBLP:conf/icml/GaborKS98}~\citep{DBLP:conf/icml/GaborKS98} has a few drawbacks, especially in the DQN setting: 
\begin{enumerate}
\setlength\itemsep{1em}
  \item Eq. \ref{Qhat_eq} is only an approximate fix point equation for the true $\hat{Q}_{i}^{*}(\mathbf{s}, a)$, because the inequality in Eq. \ref{Qhat} is arbitrarily replaced by equality.

  \item Since 
  \begin{displaymath}
  \sum_{\mathbf{s'}}{P(\mathbf{s'}|\mathbf{s},a) \max_{a' \in \{\pi_{i-1}(\mathbf{s})| \pi_{i-1} \in \Pi_{i-1}\}}{\hat{Q}_{i}^{*}(\mathbf{s'}, a' | \theta)}}
  \end{displaymath}
  is estimated by samples of $\mathbf{s'}$, and
  \begin{align}
  \label{bias}
  \begin{split}
   & E_{\mathbf{s'} \sim P(\mathbf{s'}|\mathbf{s},a)}\left[\min {\left( \tau_{i}, r_{i}(\mathbf{s},a) + 
    \gamma_{i} \max_{a' \in \{\pi_{i-1}(\mathbf{s})| \pi_{i-1} \in \Pi_{i-1}\}}{\hat{Q}_{i}^{*}(\mathbf{s'}, a' | \theta) } \right)}\right] \\
   & \leq \min {\left( \tau_{i}, E_{\mathbf{s'} \sim P(\mathbf{s'}|\mathbf{s},a)}\left[ r_{i}(\mathbf{s},a) + 
    \gamma_{i} \max_{a' \in \{\pi_{i-1}(\mathbf{s})| \pi_{i-1} \in \Pi_{i-1}\}}{\hat{Q}_{i}^{*}(\mathbf{s'}, a' | \theta)} \right] \right)}
  \end{split}
  \end{align}
  where $\theta$ is the parameter of the function approximator, the estimation is biased, similar to the bias introduced by the \texttt{max} operator in DQN as discussed in \citep{DBLP:journals/corr/HasseltGS15}.
  
  \item Noise in function approximation can create additional bias due to the \texttt{min} operator. Consider the \texttt{safety} objective where the reward is $-1$ when ego vehicle collides, $0$ otherwise. Assume that $0 \geq \tau_{i} \geq -1$, and $\mathbf{s}$ is a safe state, so that $\exists A_{s} \neq \emptyset$ s.t. $\hat{Q}_{i}^{*}(\mathbf{s}, a) = \tau_{i},  \forall  a \in A_{s}$. The target for $\hat{Q}_{i}^{*}(\mathbf{s}, a), a \in A_{s}$ computed from the right-hand-side of Eq. \ref{Qhat_eq} is
  \begin{align*}
  \begin{split}
  \min \left(\tau_{i}, r_{i}(\mathbf{s},a) +  \gamma_{i} \max_{a' \in \{\pi_{i-1}(\mathbf{s})| \pi_{i-1} \in \Pi_{i-1}\}}{\hat{Q}_{i}^{*}(\mathbf{s'}, a' | \theta)}\right) & \leq \\
  \min \left(\tau_{i}, \gamma_{i} \max_{a' \in \{\pi_{i-1}(\mathbf{s})| \pi_{i-1} \in \Pi_{i-1}\}}{\hat{Q}_{i}^{*}(\mathbf{s'}, a' | \theta)}\right)
  \end{split}
  \end{align*}
  For the target to be correct, 
  \begin{displaymath}
  \gamma_{i} \max_{a' \in \{\pi_{i-1}(\mathbf{s})| \pi_{i-1} \in \Pi_{i-1}\}}{\hat{Q}_{i}^{*}(\mathbf{s'}, a' | \theta)} \geq \tau_{i}
  \end{displaymath}
  must hold, which means that:
  \begin{align*}
  \begin{split}
  \Delta Q & = \max_{a' \in \{\pi_{i-1}(\mathbf{s})| \pi_{i-1} \in \Pi_{i-1}\}}{\hat{Q}_{i}^{*}(\mathbf{s'}, a')} - \max_{a' \in \{\pi_{i-1}(\mathbf{s})| \pi_{i-1} \in \Pi_{i-1}\}}{\hat{Q}_{i}^{*}(\mathbf{s'}, a' | \theta)} \\
           & \leq \tau_{i} - \max_{a' \in \{\pi_{i-1}(\mathbf{s})| \pi_{i-1} \in \Pi_{i-1}\}}{\hat{Q}_{i}^{*}(\mathbf{s'}, a' | \theta)} \leq  (1-\frac{1}{\gamma_{i}})\tau_{i} 
  \end{split}      
  \end{align*}
  where ${\hat{Q}_{i}^{*}(\mathbf{s'}, a')}$ is the true $\hat{Q}_{i}^{*}$ function, and $\Delta Q$ is the noise of function approximation. In other words, the noise in neural network must be smaller than $(1-\frac{1}{\gamma_{i}})\tau_{i}$ to avoid creating additional bias. If the look-ahead horizon is long, so that $\gamma_{i}  \approx 1$, the margin is very small.
  
  \item There's no guarantee the DQN will converge to the true Q value~\citep{580874}, and the learned Q function is empirically very inaccurate. Therefore, using a static threshold $\tau_{i}$ might be problematic, and an adaptive threshold that depends on the learned Q function might be preferrable.
\end{enumerate}
Observe that the only purpose of introducing $\hat{Q}_{i}^{*}(\mathbf{s}, a)$ is to bring some relaxation to 
$\max_{a \in \{\pi_{i-1}(\mathbf{s})| \pi_{i-1} \in \Pi_{i-1}\}}{Q_{i}^{*}(\mathbf{s}, a)}$
so that all actions in $\{ a  \in \{\pi_{i-1}(\mathbf{s})| \pi_{i-1} \in \Pi_{i-1}\} | Q_{i}^{*}(\mathbf{s}, a) \geq \tau_{i} \}$ are treated as equally `good enough' for that objective. So instead of estimating $\hat{Q}_{i}^{*}(\mathbf{s}, a)$, which introduces bias through the \texttt{min} operator, we can estimate $Q_{i}^{*}(\mathbf{s}, a)$ directly through the following Bellman equation:
\begin{equation}
\label{Q}
  Q_{i}^{*}(\mathbf{s}, a) = r_{i}(\mathbf{s},a) + 
    \gamma_{i} \sum_{s'}{P(\mathbf{s'}|\mathbf{s},a) \max_{a' \in \{\pi_{i-1}(\mathbf{s})| \pi_{i-1} \in \Pi_{i-1}\}}{Q_{i}^{*}(\mathbf{s'}, a')}}
\end{equation}
where $\Pi_{i}$ is \emph{redefined} as:
\begin{align}
\label{eq:policy}
\begin{split}
\Pi_{i} \stackrel{def}{=} \biggl\{ \pi_{i} \in \Pi_{i-1} \bigg| & Q_{i}^{*}(\mathbf{s}, \pi_{i}(\mathbf{s})) \geq  \max_{a \in \{\pi_{i-1}(\mathbf{s})| \pi_{i-1} \in \Pi_{i-1}\}}{Q_{i}^{*}(\mathbf{s}, a)} + \tau_{i} \\
& \mbox{ or } \pi_{i}(\mathbf{s}) = \mbox{argmax}_{a \in \{\pi_{i-1}(\mathbf{s})| \pi_{i-1} \in \Pi_{i-1}\}}{Q_{i}^{*}(\mathbf{s}, a)} \\
& ,\forall \mathbf{s} \in S \biggr\} \quad ,i = 1, 2, ..., k  
\end{split}
\end{align}
Note that the fixed threshold has been replaced by an adaptive threshold that depends on the learned Q function, and the algorithm essentially becomes the Q-learning version of \emph{lexicographic value iteration}~\citep{DBLP:conf/aaai/WrayZM15, DBLP:conf/aaaifs/PinedaWZ15}. Here, $\tau_{i}$ has a different meaning. It specifies how much worse than the best action is considered acceptable in each state. With an adaptive threshold of the form of Eq. \ref{eq:policy}, it's guaranteed that 
$\forall \pi \in \Pi_{i}, v^{\pi}_{i}(\mathbf{s}) > \max_{\pi' \in \Pi_{i-1}} v^{\pi'}_{i}(\mathbf{s}) + \frac{\tau_{i}}{1-\gamma}$.

The update rule implied by Eq. \ref{Q} for objective $i, i = 1, 2, ..., k$ is similar to Q-learning, except that the next action $a'$ is now restricted to those allowed by objective $i-1$ (In the case of $i = 1$, it degenerates to Q-learning). Once objective $i-1$ converges, it becomes regular Q-learning for an MDP whose action space is dependent on $s$. During training, one of the objectives $i  \in  \{1, 2, ..., k\}$ can be chosen for exploration at each simulation step. If objective $i$ is chosen for exploration, objectives $j = i+1, i+2, ..., k$ are no longer considered for action selection. The action selection procedure is described in algorithm \ref{alg:max_a}.

\begin{algorithm}
  \caption{Action Selection}\label{alg:max_a}
  \begin{algorithmic}[1]
  \Function{select\_action}{$Q^{*}, \mathbf{s}$}
    \Statex // $Q^{*} = [Q^{*}_{1}, Q^{*}_{2}, ..., Q^{*}_{k}]$ is the list of 
    \Statex // learned Q functions for each objective
    \State $A_{0}(\mathbf{s}) := A$    
     \For{$i$ in $\{1, 2, ..., k\}$}
       \State 
       \begin{align*}
       \begin{split}
       A_{i}(\mathbf{s}) := \biggl\{a \in A_{i-1}(\mathbf{s}) \bigg| & Q_{i}^{*}(\mathbf{s}, a) \geq \max_{a' \in A_{i-1}}{Q_{i}^{*}(\mathbf{s}, a')} + \tau_{i} \\
       & \mbox{\textbf{ or }} a = \argmax_{a' \in A_{i-1}}{Q_{i}^{*}(\mathbf{s}, a')} \biggr\}
       \end{split}
       \end{align*}
       \If{objective $i$ is chosen to be explored}
       \State \Return random action from $A_{i-1}(\mathbf{s})$
       \EndIf
     \EndFor
     \State \Return random action from $A_{k}(\mathbf{s})$
  \EndFunction
  \end{algorithmic}
\end{algorithm}

Since the only interface between objectives is the set of acceptable actions for that objective, not all objectives have to be RL agents (some of them can be rule-based agents), as long as they provide the same interface.

\subsection{Factored Q Function}
Consider the \texttt{safety} objective of self-driving, and the factored representation of state $\mathbf{s} = (s_{e}, s_{1}, s_{2}, ..., s_{m})$ , where $s_{e}$ is the state variable for ego vehicle, and $s_{1}, s_{2}, ..., s_{m}$ are the state variables for the surrounding vehicles. Informally, the problem has the following internal structure:
\begin{inparaenum}
\item collision is \emph{directly} related to only a small subset of vehicles (in most cases, ego vehicle and the vehicle ego is crashing into), so it's natural to view the reward as a function of some locally-scoped rewards $r(\mathbf{s}) = f(r(s_{e}, s_{1}), r(s_{e}, s_{2}),... , r(s_{e}, s_{m}))$
\item In some cases, $(s_{e}|_{t+1}, s_{i}|_{t+1})$ is only weakly dependent on $s_{j}|_{t}, j \neq i$, where $s_{i}|_{t}$ denotes the value of $s_{i}$ at time $t$. For example, a vehicle on the right-turn lane doesn't have much influence on the next state of a vehicle approaching the intersection from the other side.
\end{inparaenum}
Formal formulation of what it means by being \emph{`weakly'} dependent, and its effect on the value function, is difficult. However, it's reasonable to hypothesize that these structures result in some kind of structure in the value function. In fact, the task of driving safe can be thought of as the composition of a set of smaller tasks: driving safely with regard to \emph{each individual vehicle}. If we learn how to drive safely \emph{with regard to each individual vehicle}, we can use the knowledge to help with the original task of driving safely. In other words, we can use the Q functions of the smaller tasks as auxiliary features for the  Q function of the bigger original task. This idea can be formalized as follows.

Viewing $(s_{e}, s_{i}), i = 1, 2, ..., m$ as observations from the original factored MDP, and the locally-scoped rewards $r(s_{e}, s_{i})$ as rewards corresponding to the observations, we get a set of $m$ smaller auxiliary (partially observable) MDPs. To exploit the structure of the factored MDP,  the Q functions of these smaller MDPs (ignoring the partial observability) can be used as features for the Q function of the original factored MDP. To be more specific, instead of approximating the Q function of the factored MDP $Q^{*}(\mathbf{s}, a | \theta)$ directly, we learn an approximation of the Q functions of the auxiliary MDPs $Q^{*}((s_{e}, s_{i}), a | \theta_{i})$, and use these auxiliary Q functions as additional features $\phi(\mathbf{s}) = [Q^{*}((s_{e}, s_{1}), a | \theta_{1}), ..., Q^{*}((s_{e}, s_{m}), a | \theta_{m})]$ for Q function of the factored MDP.  Now the original Q function can be approximated using the augmented feature $(\mathbf{s} , \phi(\mathbf{s}))$, so we have $Q^{*}((\mathbf{s} , \phi(\mathbf{s})), a | \theta')$. The assumption here is that the additional features $\phi(\mathbf{s})$ will help with the learning of $Q^{*}(\mathbf{s},a)$. During training, these factored Q functions in $\phi(\mathbf{s})$ are updated according to their own TD errors. Section \ref{sec:network_architecture} describes this idea in the context of neural networks.

\subsection{State Space}
The state space needs to include all the necessary information for driving (vehicle kinematics, road information, etc.), and should be at such an abstraction level that policies learned on a particular road are readily transferable to roads with slightly different geometry. Our state space consists of three parts: 
\begin{enumerate}
\item ego state (table \ref{tab:state1}); 
\item state of surrounding vehicles relative to ego (table \ref{tab:state2});
\item road structure, expressed by topological relations between surrounding vehicles and ego (table \ref{tab:relation}).
\end{enumerate}
Only a subset of the state variables might be needed for each objective, e.g. the \texttt{safety} objective does not need to consider road priority information, since the goal of \texttt{safety} is to learn a generic collision avoidance policy.

\begin{table}
\caption{State Space --- Ego State}
\label{tab:state1}
\begin{center}
\begin{tabular}{ll} 
 \toprule
 $s_{e}$ & ego state \\ [0.5ex] 
 \midrule
 $v_{e}$ & ego speed \\ 
 $d_{e}$ & distance to intersection \\
 $\mathit{in\_intersection}_{e}$  & whether in intersection  \\
 $\mathit{exist\_left\_lane}_{e}$  & whether left lane exists \\
 $\mathit{exist\_right\_lane}_{e}$  & whether right lane exists\\
 $\mathit{lane\_gap}_{e}$  & lateral offset from correct (turning) lane \\
 \bottomrule
\end{tabular}
\end{center}
\end{table}

A maximum of $m$ surrounding vehicles are considered. If there are more vehicles in the scene, only the $m$ closest vehicles are considered. $\mathit{exist\_vehicle}_{1...m}$ is included in the state space in case the number of vehicle is fewer than $m$. In the experiment of this paper $m = 32$.

\begin{table}
\caption{State Space --- Surrounding Vehicles}
\label{tab:state2}
\begin{center}
 \begin{tabular}{ll} 
 \toprule
 $s_{1...m}$ & surrounding vehicles \\ [0.5ex] 
 \midrule
 $\mathit{exist\_vehicle}_{1...m}$ & whether vehicle exists \\ 
 $v_{1...m}$ & relative speed to ego \\ 
 $d_{1...m}$ & distance to intersection \\
 $\mathit{in\_intersection}_{1...m}$ & whether in intersection  \\
 $\mathit{exist\_left\_lane}_{1...m}$  & whether left lane exists \\
 $\mathit{exist\_right\_lane}_{1...m}$  & whether right lane exists\\
 $x_{1...m}$ , $y_{1...m}$ & relative position to ego \\
 $\theta_{1...m}$ & relative heading to ego\\
 $\mathit{has\_priority}_{1...m}$ & whether has right-of-way over ego \\
 $\mathit{ttc}_{1...m}$ & time-to-collision with ego \\
 $\mathit{signal}_{1...m}$ & brake/turn signal \\
 \bottomrule
\end{tabular}
\end{center}
\end{table}

\begin{figure}
\centering
\includegraphics[width=2in]{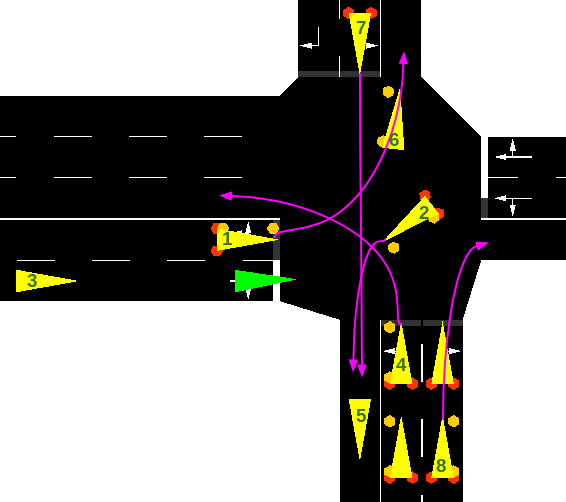}
\caption{Illustration of topological relations. With respect to the green vehicle, vehicle $2,4,7$ are $\mathit{crossing}$, vehicle $8$ is $\mathit{merge}$, vehicle $5$ and $6$ are $\mathit{irrelevant}$. Vehicle $1$ is to the $\mathit{left}$ of the green vehicle, and the latter is to the $\mathit{right}$ of the former. Vehicle $3$ is $\mathit{behind}$ the green vehicle, and the latter is $\mathit{ahead}$ of the former.}
\label{fig:intersection}
\end{figure}

In order to deal with complex roads with multiple lanes, topological
relations between ego and each surrounding vehicle also need to be included. Inspired by the lanelet model introduced by \citeauthor{DBLP:conf/ivs/BenderZS14}~\citep{DBLP:conf/ivs/BenderZS14}, we define seven topological relations between vehicles (table \ref{tab:relation}), which are illustrated in figure \ref{fig:intersection}. These relations capture the interconnection between roads through vehicles in the scene and their intended path, without explicitly modelling the road structure.

\begin{table}
\caption{State Space --- Topological Relations with Ego}
\label{tab:relation}
\begin{center}
 \begin{tabular}{ll} 
 \toprule
 $\mathit{merge}$ & merging into the same lane \\ 
 $\mathit{crossing}$ & routes intersecting each other\\ 
 $\mathit{left}$ & in left lane\\ 
 $\mathit{right}$ & in right lane\\ 
 $\mathit{ahead}$ & ahead in the same or succeeding lane\\ 
 $\mathit{behind}$ & behind in the same or previous lane\\ 
 $\mathit{irrelevant}$ & none of the above\\ 
 \bottomrule
\end{tabular}
\end{center}
\end{table}

\subsection{Network Architecture} \label{sec:network_architecture}
\begin{figure}
\centering
\includegraphics[width=1.75in]{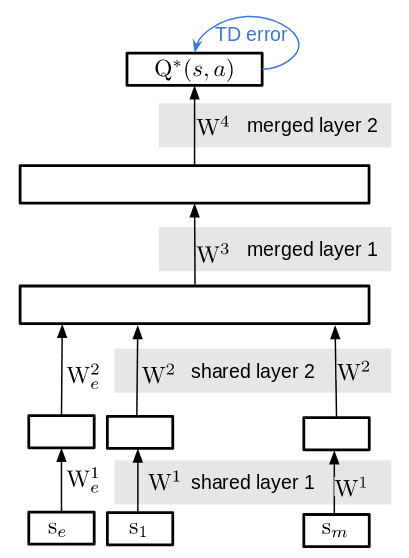}
\caption{Neural network architecture with built-in invariance to the order of surrounding vehicles in the state. We first pass $s_{1}, s_{2}, ..., s_{m}$ through a few shared layers to get the corresponding features. Then these features are merged through addition and activation. After that, the network is fully-connected.}
\label{fig:nn1}
\end{figure}

The state $s = (s_{e}, s_{1}, s_{2}, ..., s_{m}) \in S$ contains the state variables of $m$ surrounding vehicles $s_{i}, i = 1, 2, ..., m$ (including their topological relations with ego). Since swapping the order of two surrounding vehicles in the state doesn't change the scene, the Q value should remain the same: 
\begin{displaymath}
Q((s_{e}, s_{1}, ...,  s_{i}^{u}, ..., s_{j}^{v}, ..., s_{m}), a) = Q((s_{e}, s_{1}, ..., s_{j}^{v}, ..., s_{i}^{u}, ..., s_{m}), a)
\end{displaymath}
where $s_{i}^{u}$ denotes the $u$th possible instantiation of $\dom (s_{i})$. To build this invariance into the neural network, the network needs to be symmetric with respect to each $s_{i}$. In other words, the weights connecting $Q(s,a)$ to each $s_{i}$ should be the same (shown in Figure \ref{fig:nn1}).

\begin{figure}
\centering
\includegraphics[width=1.75in]{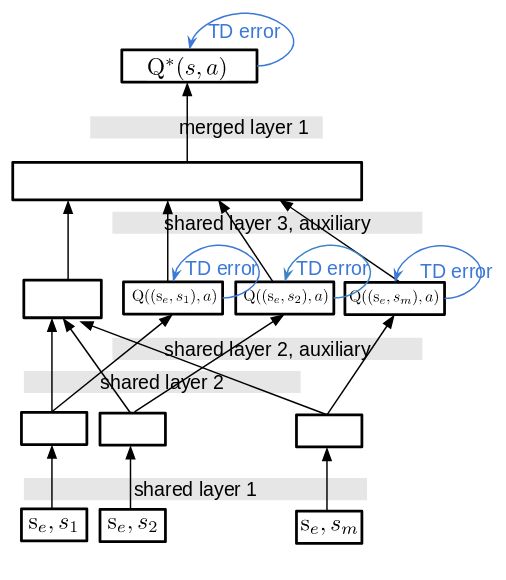}
\caption{Neural network architecture with factored Q function. The shared layer branches off for the factored Q functions (auxiliary branch), which is merged back in higher layers.}
\label{fig:nn2}
\end{figure}

If factored Q function is used, then $m$ additional heads for these value functions are needed (Figure \ref{fig:nn2}). During each update, $m$ Q functions are improved simultaneously in addition to the original Q function, each of which corresponds to learning to avoid collision with each of the $m$ surrounding vehicles, in the case of the \texttt{safety} objective. Since the agent utilizes a single scene to learn multiple aspects within the scene, better data efficiency can be expected.

\section{Experiment}
SUMO (Simulation of Urban Mobility~\citep{DBLP:conf/sumo/2014}) traffic simulator is used as the simulation environment for our experiment. We wrote a RL interface similar to OpenAI Gym on top SUMO to provide the state and action space. ~\footnote{Source code can be found at \url{https://gitlab.com/sumo-rl/sumo_openai_gym}} Given a map, the set of all possible routes a vehicle can travel is predefined. The vehicle needs to control throttle and lane change behavior. The action space is a discrete set of $9$ actions: 
\begin{multicols}{2}
\begin{enumerate}
    \item \texttt{max\_deceleration};
    \item \texttt{med\_deceleration};
    \item \texttt{min\_deceleration};
    \item \texttt{maintain\_speed};
    \item \texttt{min\_acceleration};
    \item \texttt{med\_acceleration};
    \item \texttt{max\_acceleration};
    \item \texttt{change\_to\_right\_lane};
    \item \texttt{change\_to\_left\_lane}
\end{enumerate}
\end{multicols}
The vehicle kinematics follows a point-mass model, and lane changes are instantaneous. Speed is assumed to be maintained during lane changes. The learning scenario is a typical urban four-way intersection of a major road and a minor road with random traffic. Traffic coming from the minor road needs to yield to the major road, and turns need to be made from the correct lane(s).

\subsection{Objectives}
We consider four objectives in this paper. In lexicographic order, the objectives are: 
\begin{enumerate}
\item \texttt{lane_change}: rule-based; all it does is to rule out invalid lane change actions, namely: lane change to the left/right when there's no left/right lane, and lane change in intersections. The state space only has three state variables: $\mathit{exist\_left\_lane}_{e}$, $\mathit{exist\_right\_lane}_{e}$ and $\mathit{in\_intersection}_{e}$; thus, it's trivial to learn even if it were implemented as a RL agent.
\item \texttt{safety}: RL-based; it ensures that collision doesn't happen. $-1$ reward if collision occurs, or if time-to-collision with \emph{at least one} surrounding vehicle is less than $3$s and is still decreasing; $0$ reward otherwise.~\footnote{This is only a simplified description of the actual reward used. Since we use a simple calculation for time-to-collision, sometimes it's not suitable to make the reward dependent on the (inaccurate) estimates of time-to-collision. In these cases, the reward is set to $0$. For the intricacies of the reward function, please refer to the source code.} The state space includes everything except $\mathit{lane\_gap}_{e}$ and $\mathit{has\_priority}_{1..m}$.  Factored Q functions are learned on the auxiliary MDPs
\begin{displaymath}
\left( \dom(s_{e}, s_{i}), A, \gamma, r_{i} \right) , i = 1, 2, ..., m
\end{displaymath}
Where $r_{i}$ is just the locally-scoped version of $r$: $-1$ if ego collides with vehicle $i$ or the time-to-collision with \emph{vehicle $i$} is less than $3$s and is still decreasing; $0$ reward otherwise.
Since up to $m = 32$ vehicles are considered, up to $32$ instances of auxiliary POMDPs (which share the state space with the original factored MDP) can be running at the same time. If vehicle $i$ goes out of scene or crashes with ego vehicle, the episode ends for instance $i$ of the auxiliary task. Adaptive threshold is used, and $\tau$ is set to $-0.2$ during training; then it's manually fine-tuned on the training set before testing.
\item \texttt{regulation}: RL-based; it makes sure that traffic rules are followed. We consider two traffic rules: 
\begin{inparaenum}
\item to make turns from the correct lane(s);
\item to yield according to right-of-way.
\end{inparaenum}
A reward of $-1$ is given for failure to yield right-of-way, $-0.02$ for failure to proceed when having right-of-way, and up to $-1$ for staying in the wrong lane (e.g. staying in the left-turn lane, if the assigned route is straight). The state space is comprised of $\mathit{has\_priority}_{1...m}$, $\mathit{lane\_gap}_{e}$, $\mathit{in\_intersection}_{e}$, $v_{e}$ and $d_{e}$. Change of right-of-way or change of road is considered end of episode, since these changes would happen regardless of the actions chosen. $\tau$ is set to $-0.2$ during training.
\item \texttt{comfort\&speed}: rule-based; prefers acceleration unless speed limit is reached, while avoiding extreme actions (e.g. maximum acceleration) and lane changes.
\end{enumerate}

\subsection{Training}
The agent is trained on two intersecting roads with random surrounding traffic. Traffic enters the scene with a random probability in each episode. An episode ends either when ego collides with other vehicle(s) or when the timeout is reached. Each surrounding vehicle has a normally distributed maximum speed, and is controlled by SUMO's rule-based behavioral model, which attempts to mimic human drivers. The intersection part of the map is shown in Figure \ref{fig:intersection}. The north/south-bound traffic needs to yield to the east/west-bound traffic. In each episode, ego vehicle is randomly assigned one of the possible routes within the map.  Each RL-based objective is trained using double DQN~\citep{DBLP:journals/corr/HasseltGS15} with prioritized experience replay~\citep{DBLP:journals/corr/SchaulQAS15}. To speed up training, $10$ simulation instances run in parallel, adding experience to the experience replay buffer. Asynchronous~\citep{DBLP:conf/icml/MnihBMGLHSK16} update is performed on the Q functions of each objective. 

Three models are trained for comparison, which we later refer to as \textbf{DQN},  \textbf{TLDQN}, and \textbf{TLfDQN} respectively: 
\begin{enumerate}
\item Scalar-valued DQN: The neural network architecture is as shown in Figure \ref{fig:nn1}, with $4$ shared layers and $2$ merged layers. Each layer has $64$ hidden units. The reward function is a weighted sum of the rewards used for the multi-objective case. The weights are chosen in a way that try to reflect the relative importance of each objective.
\item Thresholeded lexicographic DQN: The \texttt{safety} objective uses the same neural network architecture as above. The \texttt{regulation} objective uses a $4$-layer fully connected network with $64$ hidden units in each layer.
\item Thresholded lexicographic DQN with factored Q function: The \texttt{safety} objective uses the neural network architecture as shown in Figure \ref{fig:nn2}, but with only the auxiliary branch. The auxiliary branch has $4$ shared layers, each with $64$ hidden units; the merged layer is a fixed \texttt{min} layer that takes the minimum of the factored Q functions for each action.
$Q(\mathbf{s}, a | \theta) = \min_{i} Q((s_{e}, s_{i}), a | \theta)$
The \texttt{regulation} objective uses the same network structure as above. 
\end{enumerate}

\subsection{Results}
\begin{figure*}
    \centering
    \subfloat[slow down]{\includegraphics[width=1in]{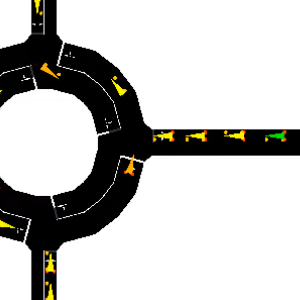}}
    \hspace{.25in}
    \subfloat[yield]{\includegraphics[width=1in]{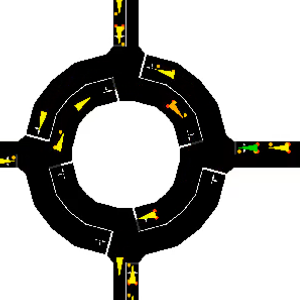}}
    \hspace{.25in}
    \subfloat[on ring road]{\includegraphics[width=1in]{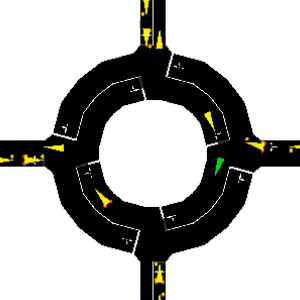}}
    \hspace{.25in}
    \subfloat[exit]{\includegraphics[width=1in]{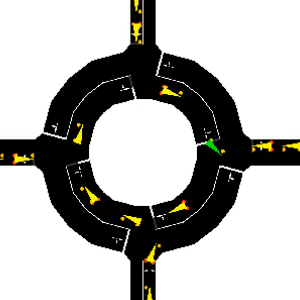}}
    \caption{Zero-shot transfer performance. The agent (green) is able to drive through the ring road without safety or traffic rule violation.}
    \label{fig:ringroad}
\end{figure*}

The three models are first evaluated on the same intersecting roads they've been trained on, with random traffic; then their zero-shot transfer performance is evaluated on a ring road (Figure \ref{fig:ringroad}) they've never seen during training. The vehicle can enter the ring road through either right or left turn. Traffic entering the ring road needs to yield to traffic already on the ring road.

\begin{figure}
\centering
\includegraphics[width=2.5in]{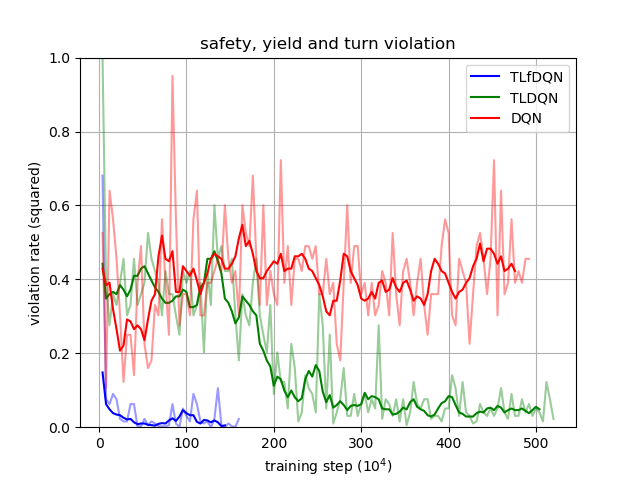}
\caption{Learning curve of DQN, TLDQN and TLfDQN. The dark curves are the moving averages.}
\label{fig:curve}
\end{figure}

Figure \ref{fig:curve} shows the learning curve of DQN, TLDQN and TLfDQN. The x-axis is the training step, and the y-axis is the (squared) rate of safety (collisions) and traffic rule (yielding and turning) violations combined. Timeouts are counted as yielding violations. TLDQN and DQN are trained for $30$ hours, while TLfDQN is only trained for $18$ hours since it has already converged. We see that TLfDQN is able to reach a good policy within $500, 000$ training steps, as compared to $3$ million training steps for TLDQN, improving the data efficiency by $6$ times. It should be noted that the training time of TLfDQN per training step is longer than TLDQN ($26$ minutes as compared to $14$ minutes), mostly due to the computational overhead of the $32$ additional targets for the factored Q functions, one for each surrounding vehicle in the scene. However, the overhead can potentially be alleviated by parallelizing the computation of the target. Within $30$ hours of training, scalar-valued DQN is not able to learn an acceptable policy, indicating the effectiveness of the multi-objective approach. Different weightings for the objectives in the reward function were tried for scalar-valued DQN, no significantly better result was observed.~\footnote{A good weighting scheme for the reward might exist, but nevertheless hard to find; and to test a set of new weights, the agent has to be re-trained.}

\begin{figure*}
    \centering
    \subfloat[DQN]{\includegraphics[width=2.5in]{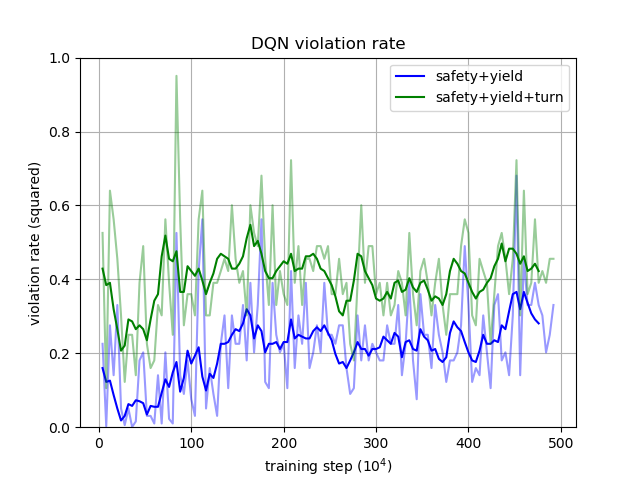}}
    \subfloat[TLDQN]{\includegraphics[width=2.5in]{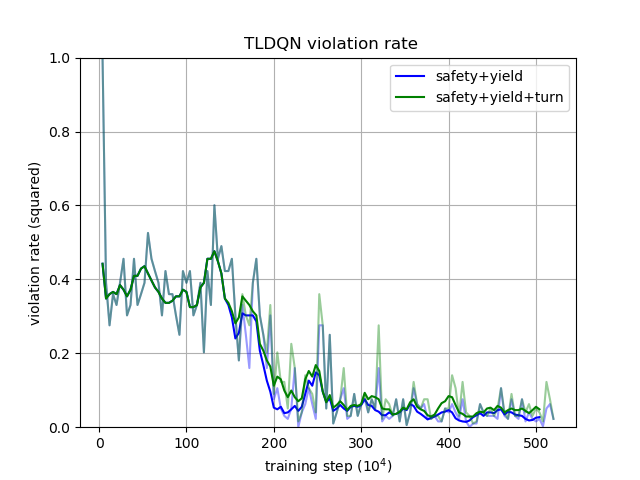}}
    \subfloat[TLfDQN]{\includegraphics[width=2.5in]{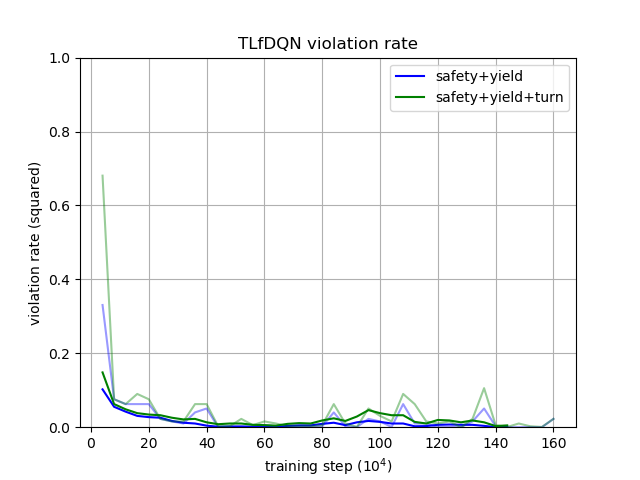}}
    \caption{Learning curves showing different types of violations. The blue curves show collisions, timeouts and failures to yield combined; the green curves show collisions, timeouts, failures to yield and failures to change to correct lane for turning combined}
    \label{fig:breakdown}
\end{figure*}

Figure \ref{fig:breakdown} shows the learning curves with a breakdown of different types of violation. Ideally, we would like to show how the agent performs on each objective. However, many violations are inter-correlated, e.g.,  safety violations are usually preceded by failure to yield; improperly stopping in the middle of the road leads to low safety violation rate; high safety violation rate often leads to lower turning violation rate, because the agent simply collides before even reaching the intersection. Therefore, we group the more serious violations --- safety, failure to yield and timeouts, into one category; and the less serious violation --- failure to change to correct lane, into another category. The blue curves show the first category, and the green curves show both categories. Note that failure to change to correct lane doesn't necessary imply a bad policy, because in some scenarios, the road is just too crowded for lane changes. We see in the figure that in both categories, TLfDQN performs the best. It's worth noting that it might seem that the scalar-valued DQN briefly achieves better performance before getting worse. However, the videos indicate that the lower collision rate is due to the agent learning an incorrect policy that stops abruptly in the middle of the road and waits until all the traffic clears before moving.

\begin{figure*}
    \centering
    \subfloat[need to change to left-turn lane]{\includegraphics[width=1in]{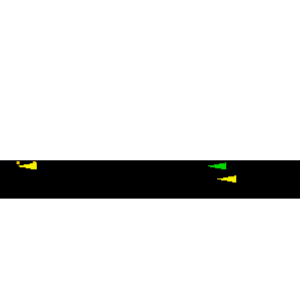}}
    \hspace{.25in}
    \subfloat[slow down to find a gap]{\includegraphics[width=1in]{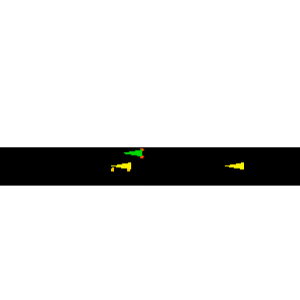}}
    \hspace{.25in}
    \subfloat[successful lane change]{\includegraphics[width=1in]{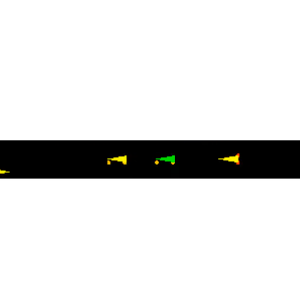}}
    \hspace{.25in}
    \subfloat[wait for traffic ahead]{\includegraphics[width=1in]{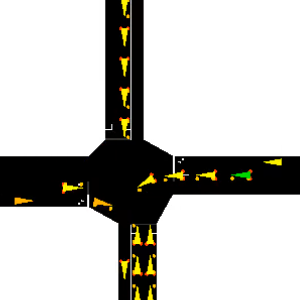}}
    \caption{Ego vehicle (green) is assigned a left-turning route in this episode. The agent slows down to find a gap and successfully changes to the left-turn lane.}
    \label{fig:lane_change}
\end{figure*}

\begin{figure*}
    \centering
    \subfloat[slow down]{\includegraphics[width=1in]{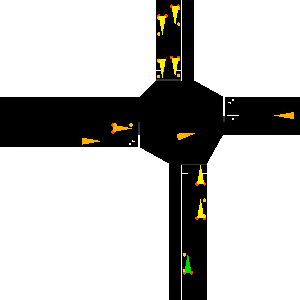}}
    \hspace{.25in}
    \subfloat[yield]{\includegraphics[width=1in]{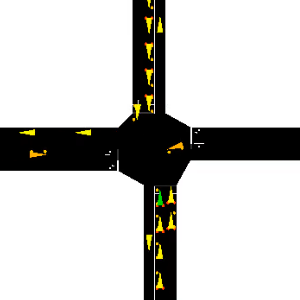}}
    \hspace{.25in}
    \subfloat[proceed when the way is clear]{\includegraphics[width=1in]{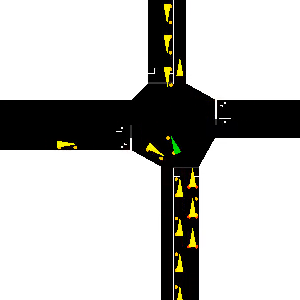}}
    \hspace{.25in}
    \subfloat[successful left turn]{\includegraphics[width=1in]{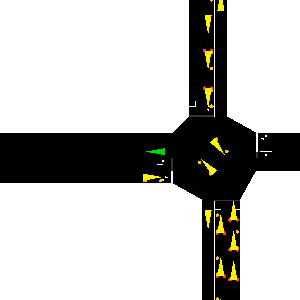}}
    \caption{The agent (green) yields to the traffic on the main road, and then proceeds when it has right-of-way.}
    \label{fig:left_turn}
\end{figure*}

Videos of the learned policy of our multi-objective RL agent can be found online~\footnote{\url{https://www.youtube.com/playlist?list=PLiZsfe-Hr4k9VPiX0tfoNoHHDUE2MDPuQ}}. Figure \ref{fig:lane_change} and Figure \ref{fig:left_turn} are some snapshots of the videos. Ego vehicle is colored as green, and vehicles that have right-of-way over ego are colored as orange. In Figure \ref{fig:lane_change}, the ego vehicle is assigned a left-turning route, so it needs to first change to the left lane, then take a left turn. The ego vehicle learns to slow down (notice the braking lights) until a gap is found, and then change lane to the left. In Figure \ref{fig:left_turn}, the ego vehicle slows down to yield for traffic on the major road, then proceeds to complete the left turn after the road is clear. The vehicle is not yielding for the right-turning vehicle because there's no conflict between them. Figure \ref{fig:ringroad} shows the zero-shot transfer performance on a ring road. The agent is able to drive through the ring road safely and yield to traffic already on the ring road before entering. The performance of the three models after $30$ hours of training evaluated on $1,000$ random episodes is shown in Table \ref{tab:result}. 

\begin{table}
\caption{Violation Rate after $30$ Hours of Training}
\label{tab:result}
\begin{tabular}{llll} 
 \toprule
 Model & Collision & Yielding & Turning\\ [0.5ex] 
 \midrule
 DQN  & $32.9\%$ & $8.5\%$ & $16.4\%$\\ 
 TLDQN & $10.9\%$ & $0.9\%$ & $7.6\%$\\ 
 TLfDQN & $3.6\%$ & $1.0\%$ & $2.4\%$\\ 
 TLfDQN (transfer) & $3.5\%$ & $0.4\%$ & N/A\\ 
 \bottomrule
\end{tabular}
\end{table}

\section{Conclusions}
We've shown in the context of autonomous urban driving that a multi-objective RL approach --- thresholded lexicographic DQN might be effective for problems whose objectives are difficult to express using a scalar reward function due to the many aspects involved. The proposed method of learning factored Q functions, and using them as auxiliary features during training is shown to improve data efficiency. Combining these ideas, we trained an autonomous driving agent that is, most of the time, able to safely drive on busy roads and intersections, while following traffic rules.

\begin{acks}
The authors would like to thank Jae Young Lee, Sean Sedwards, Atrisha Sarkar, Aravind Balakrishnan, Ashish Gaurav and Piyush Agarwal for discussions.
\end{acks}


\bibliographystyle{ACM-Reference-Format}  
\balance  
\bibliography{sample-aamas19}  

\end{document}